\documentclass{article}
\usepackage{ijcai21}

\usepackage{times}
\usepackage{float} 
\usepackage{subfigure}
\usepackage{soul}
\usepackage{url}
\usepackage[hidelinks]{hyperref}
\usepackage[utf8]{inputenc}
\usepackage[small]{caption}
\usepackage{graphicx}
\usepackage{amsmath}
\usepackage{amsthm}
\usepackage{amsfonts}
\usepackage{booktabs}
\usepackage{algorithm}
\usepackage{algorithmic}
\usepackage{amsmath}
\usepackage{bbm}
\usepackage{bm}
\usepackage{booktabs}
\usepackage{dsfont}

\usepackage[dvipsnames]{xcolor}

\title{Multi-Scale Contrastive Siamese Networks for Self-Supervised Graph Representation Learning}

\author{
Ming Jin$^1$\and
Yizhen Zheng$^1$\and
Yuan-Fang Li$^1$\and
Chen Gong$^2$\and
Chuan Zhou$^3$\And
Shirui Pan$^1$\footnote{Corresponding author}\\
\affiliations
$^1$Department of Data Science and AI, Faculty of IT, Monash University, Australia\\
$^2$School of Computer Science and Engineering, Nanjing University of Science and Technology, China\\
$^3$Academy of Mathematics and Systems Science, Chinese Academy of Sciences, China\\
\emails
\{ming.jin, yizhen.zheng, yuanfang.li, shirui.pan\}@monash.edu,
chen.gong@njust.edu.cn,
zhouchuan@amss.ac.cn
}

\begin{document}

\maketitle

\begin{abstract}
Graph representation learning plays a vital role in processing graph-structured data. However, prior arts on graph representation learning heavily rely on labeling information. To overcome this problem, inspired by the recent success of graph contrastive learning and Siamese networks in visual representation learning, we propose a novel self-supervised approach in this paper to learn node representations by enhancing Siamese self-distillation with multi-scale contrastive learning. Specifically, we first generate two augmented views from the input graph based on local and global perspectives. Then, we employ two objectives called cross-view and cross-network contrastiveness to maximize the agreement between node representations across different views and networks. To demonstrate the effectiveness of our approach, we perform empirical experiments on five real-world datasets. Our method not only achieves new state-of-the-art results but also surpasses some semi-supervised counterparts by large margins.
Code is made available at \href{https://github.com/GRAND-Lab/MERIT}{https://github.com/GRAND-Lab/MERIT}
\end{abstract}

\section{Introduction}
Over the past few years, 
graph representation learning, which aims to learn low-dimensional embeddings of nodes or graphs to preserve the underlying structural and attributive information, has become a pivotal part of mining graph-structured data. The learned embeddings can then be used in different downstream tasks, such as node and graph classification, by training specific 
decoders on top of the learned embeddings. Although 
graph neural networks (GNNs) 
~\cite{chebyshev,gcn,gat,sage} have achieved significant progress in graph representation learning, most of them require a certain number of labeled nodes to train, which hinders them from being adopted in real-world applications where the labeling information is usually scarce and valuable.

To mitigate this gap, self-supervised graph representation learning approaches, especially those methods based on contrastive learning,  have recently achieved promising results. Traditional unsupervised methods such as DeepWalk~\cite{deepwalk} and Node2Vec~\cite{node2vec} are based on random walks and the skip-gram model, forcing neighboring nodes to have similar representations. However, random walk and other matrix reconstruction-based approaches~\cite{vgae,sage} place a strong emphasis on the graph proximity and fail to consider other widely available relationships within or between subgraphs. Following the concept of mutual information (MI)~\cite{cpc} and other visual representation learning advances~\cite{cmc,moco,simclr,dim}, a series of graph contrastive learning (GCL) methods have been proposed. For example, inspired by Deep InfoMax~\cite{dim}, DGI~\cite{dgi} proposes to maximize the mutual information between the patch- and global-level representations. Based on this, MVGRL~\cite{mvgrl} introduces the concept of graph multi-view contrastive learning by discriminating the patch-global representations over two augmented views that derived from the input graph. Other approaches, such as GMI~\cite{gmi} and GRACE~\cite{grace}, extend the idea of MI maximization to contrast the representation of a node with its raw information (e.g.,\ node features) or neighbors' representations in different views.

Although the aforementioned methods have achieved significant success, they  suffer all or at least partially  the following limitations. Firstly, existing MI-based methods, such DGI, GMI, and MVGRL, usually require an additional  MI estimator to score positive (e.g.,\ local-global representations) and negative pairs (e.g.,\ representations from corrupted views), which is computationally expensive and also makes the model sensitive to the choice of discriminators~\cite{tschannen2019mutual}. Secondly, most of existing GCL methods heavily rely on a large number of negative samples to avoid collapsing to trivial solutions. In other words, negative nodes and graphs act as an indispensable regulator that needs to be deliberately selected in contrastive learning. To alleviate this problem, Grill et al.~\shortcite{byol} propose the Bootstrap Your Own Latent (BYOL) framework to perform unsupervised representation learning on images by leveraging the bootstrapping mechanism without using negative samples. Self-supervised representation learning methods utilizing Siamese networks~\cite{simsiam,raft} predominantly work in the visual domain but have not been extended to graphs yet.

In this paper, to alleviate the drawbacks of existing GCL methods and take advantage of bootstrapping in Siamese networks, we propose a simple yet powerful framework to learn node-level representations, which we refer to as \underline{M}ulti-scal\underline{E} cont\underline{R}astive s\underline{I}amese ne\underline{T}work (MERIT). Our method is designed to optimize two objectives, namely cross-network and cross-view contrastiveness. Firstly, considering that existing GCL methods heavily rely on negative samples to avoid representation collapse, we propose to use a momentum-driven Siamese architecture as our backbone to maximize the similarity between node representations in different views from the online and target network, respectively. The intuition behind is that the slowly-moving target network in our framework serves as a stable ``mean teacher'' to encode historical observations, which guides the online network to learn to explore richer and better representations without relying on extra negatives to avoid collapse~\cite{byol}. However, merely optimizing this objective ignores the rich underlying graph topological information. To  partially alleviate this problem, we inject additional negative samples to push disparate nodes away in different views across two networks, as shown in Figure \ref{fig:picture002}. Secondly, different from the work in the visual domain where the similarity measurements are typically defined on the image level, we propose to further utilize node connectivity and introduce a multi-scale contrastive learning within and across views in the online network, i.e., Figure \ref{fig:picture003}, to regularize the training of the aforementioned bootstrapping objective in our method. Experimental results on a variety of datasets demonstrate the superb performance of our design.

Our contribution is summarized as follow:
\begin{itemize}
  \item We propose a novel framework to learn node representation by taking advantage of bootstrapping in Siamese network and multi-scale graph contrastive learning. To the best of our knowledge, we are the first to use the Siamese networks on node representation learning.

  \item We propose two types of contrastive objectives for self-supervised node representation learning based on Siamese networks. They regularize each other and are capable of providing a more effective graph encoder.

  \item We conduct extensive experiments on various real-world datasets and validate the effectiveness of the proposed method over state-of-the-art methods in self-supervised graph representation learning.
\end{itemize}

\section{Related Work}
\begin{figure*}[t]
\centering
\includegraphics[width=12cm]{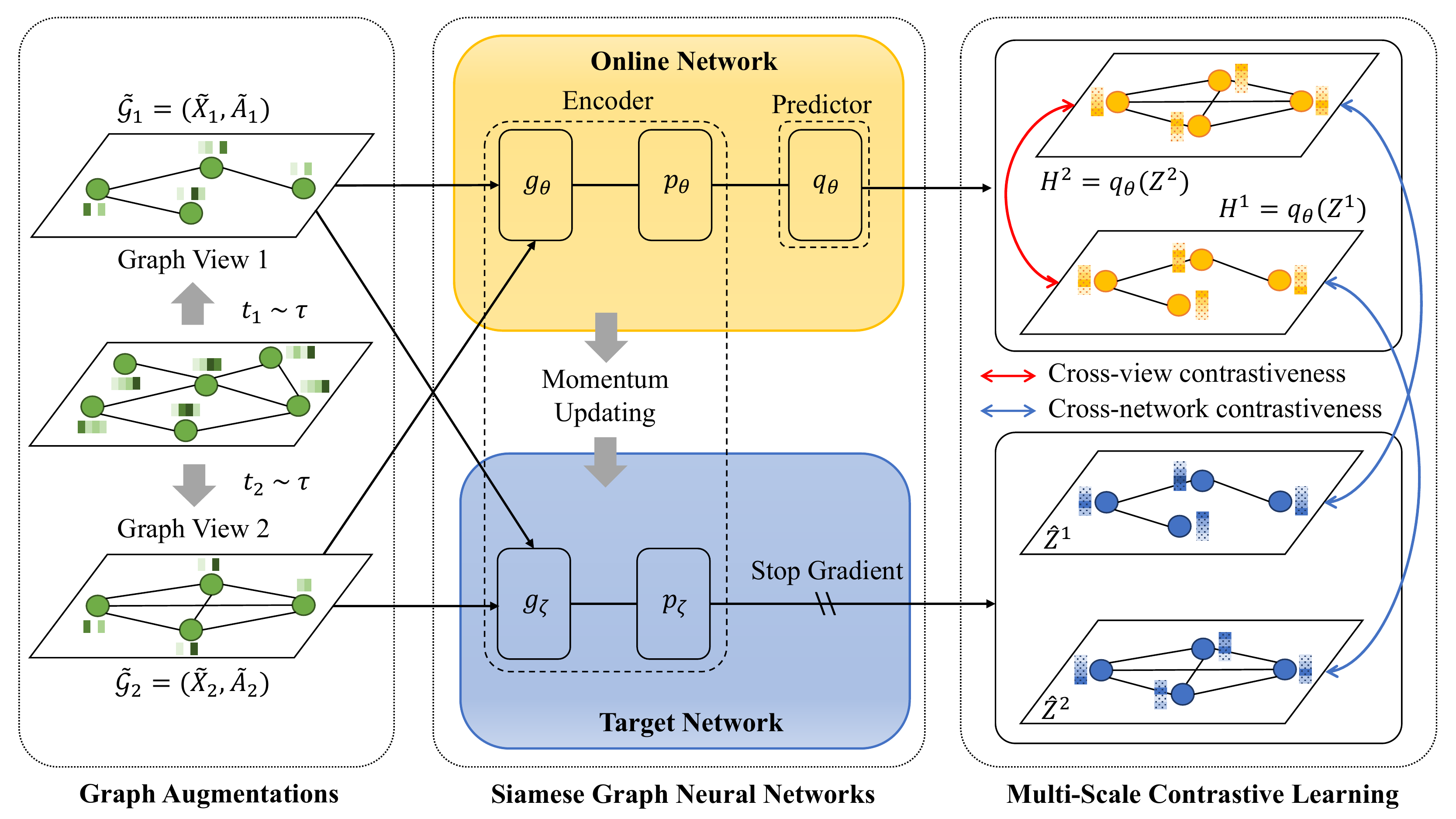}
\caption{The overall framework of MERIT. Through graph augmentations, we construct two graph views, based on which an online network and a target network are employed to generate node representations for each view. A multi-scale contrastive learning scheme, which utilizes both cross-network and cross-view contrastive modules, is deployed to learn effective node embeddings. $g_{\theta}$ and $g_{\zeta}$ denotes a GNN-based graph encoder. $p_{\theta}$, $p_{\zeta}$, and $q_{\theta}$ are two-layer MLP with the batch normalization. $t_1 \sim \tau$ and $t_2 \sim \tau$ are two different graph augmentations}
\label{fig:picture001}
\end{figure*}

\vspace{0.8mm}\noindent\textbf{Siamese network} 
is a neural architecture that contains two or more identical structures (e.g.,\ online and target encoder in Figure~\ref{fig:picture001}) to make multi-class prediction or entity comparison~\cite{bromley1993signature}. Traditionally, it has been used on  supervised tasks such as signature verification~\cite{bromley1993signature} and face matching~\cite{taigman2014deepface}.
Recently, Grill et al.~\shortcite{byol} employed this architecture on self-supervised visual representation learning and achieved significant improvements over existing arts without using negative samples. To fully understand the underlying mechanism of BYOL, SimSiam~\cite{simsiam} and RAFT~\cite{raft} verify that the extra predictor in the online network and the stop-gradient mechanism in the target network are keys to prevent the collapse without the help of negative samples.

\vspace{1mm}\noindent\textbf{Unsupervised graph representation learning} 
approaches are traditional based on random walks~\cite{deepwalk,node2vec} and adjacency matrix reconstruction~\cite{vgae}. These methods heavily rely on node proximity but are less scalable and error-prone without extracting other widely available self-supervision signals in graphs. Recently, unsupervised GNN-based methods, such as GraphSAGE~\cite{sage}, have achieved considerable progress but still limited in performance. Contrastive methods, on the other hand, alleviate the aforementioned problems and achieve state-of-the-art performance. For example, DGI~\cite{dgi} and InfoGraph~\cite{infograph} employ the idea of Deep InfoMax~\cite{dim} and consider both patch and global information during the discrimination. MVGRL~\cite{mvgrl} introduces augmented views to graph contrastive learning and optimizes the DGI-like objectives. GMI~\cite{gmi} proposes considering both graph proximity and feature space similarity by maximizing the mutual information between the representation and feature space with learnable weights. Other approaches, such as CG3~\cite{cg3} and GRACE~\cite{grace}, further extend the idea of graph MI maximization and conduct the discrimination on different scales.

\section{Proposed Method}
\paragraph{Problem definition.} 
Given a graph $\mathcal{G}=(X, A)$, where $X \in \mathbb{R}^{N \times D}$  denotes the node feature matrix, and  $A \in \mathbb{R}^{N \times N}$ indicates the adjacency matrix where each entry $A_{ij}$ is the linkage relation between nodes $i$ and $j$. In this paper, we aim to learn a graph encoder $g_{\theta}: \mathbb{R}^{N \times D} \times \mathbb{R}^{N \times N} \to \mathbb{R}^{N \times D'}$ such that $D'\ll D$, without relying on the labeling information. The resulted representations $H=g_{\theta}(X,A)=\{h_1,h_2...,h_N\}$ can then be directly used in downstream tasks, such as node classification.

\paragraph{Overall framework.} 
We propose a novel algorithm, namely MERIT, to learn node representations by taking advantage of both bootstrapping and multi-scale graph contrastive learning. As illustrated in Figure \ref{fig:picture001}, our model mainly consists of three components: Graph augmentations, cross-network contrastive learning, and cross-view contrastive learning. To train our model, we first generate two augmented graph views, denoted as $\Tilde{\mathcal{G}_1}$ and $\tilde{\mathcal{G}_2}$. After this, by processing these two views via the \textit{online network} and the \textit{target network}, we construct different graph contrastive paths on multiple scales in the latent space, as shown on the rightmost part in Figure \ref{fig:picture001}. In the following sections, we illustrate the aforementioned crucial components.

\subsection{Graph Augmentations}
Augmentation is a key component in self-supervised visual representation learning.
However, image augmentations such as cropping and rotating cannot be directly applied to graphs due to the huge disparity of these two modalities. Therefore, to facilitate contrastive learning on graphs, we propose four augmentation methods, as shown below, to augment the graph topological and attributive information.

\paragraph{Graph Diffusion (GD).} 
We transform a graph via diffusion to generate a congruent view.  
The effectiveness of this method may be attributed to the extra global information provided by the diffused view. This process is fomulated as:
\begin{equation}
S = \sum^{\infty}_{k = 0}\theta_{k}T^{k} \in \mathbb{R}^{N \times N},
\label{eq: diff}
\end{equation}
where $\theta$ is a parameter to control the distribution of local and global signals, $T \in \mathbb{R}^{N \times N}$ is the transition matrix to transfer the adjacency matrix \cite{klicpera2019diffusion}.
In this paper, we adopt the Personalized PageRank (PPR) kernel to power the graph diffusion. Formally, given the adjacency matrix $A$, the identity matrix $I$, and the degree matrix $D$, Equation~\ref{eq: diff} can be reformulated as:
\begin{equation}
S = \alpha\left(I - (1 - \alpha)D^{-1/2}AD^{-1/2}\right)^{-1},
\label{eq: ppr}
\end{equation}
where $\alpha$ is a tunable parameter for the random walk teleport probability \cite{mvgrl}.

\paragraph{Edge Modification (EM).} 
Instead of merely dropping edges in the adjacency matrix, we also add the same number of dropped edges~\cite{you2020graph}. In such a way, we can maintain the original graph's property, while complicate the augmented view with the additional edges. Specifically, given the adjacency matrix $A$ and the modification ratio $P$, we randomly drop $P/2$ portion of existing edges in the original graph and then randomly add the same portion of new edges to the graph. Both our edge dropping and adding process follow an i.i.d. uniform distribution.

\paragraph{Subsampling (SS).} 
Similar to the image cropping, we randomly select a node index in the adjacency matrix as the splitting point, and then use it to crop the original graph to create a fixed-sized subgraph as the augmented graph view. 
An advantage of SS is enabling the batch processing 
to handle large graphs whose size may exceed the capability of the GPU memory.

\paragraph{Node Feature Masking (NFM).}
Different from~\cite{you2020graph}, given the feature matrix $X$ and an augmentation ratio $P$, we randomly select $P$ fraction of node feature dimensions in $X$ and then mask them with zeros.

\vspace{1mm}In this paper, we apply SS, EM, and NFM to the first view, and use SS + GD + NFM for the second congruent view. By doing so, our model can encode both the local and global information through the contrastive learning.

\subsection{Cross-Network Contrastive Learning}
In MERIT, we introduce a Siamese architecture, which consists of two identical encoders (i.e., $g_{\theta}$, $p_{\theta}$, $g_{\zeta}$, and $p_{\zeta}$) with an extra predictor $q_{\theta}$ on top of the online encoder, as shown in Figure \ref{fig:picture001}. 
We first take node representations from one view in the online network as the anchor, then we maximize the cosine similarity to the corresponding representations from another view in the target network to form the basic bootstrapping contrastiveness. 

\begin{figure}[t]
\centering
\subfigure[Cross-network contrastiveness.]{
\includegraphics[height = 3cm]{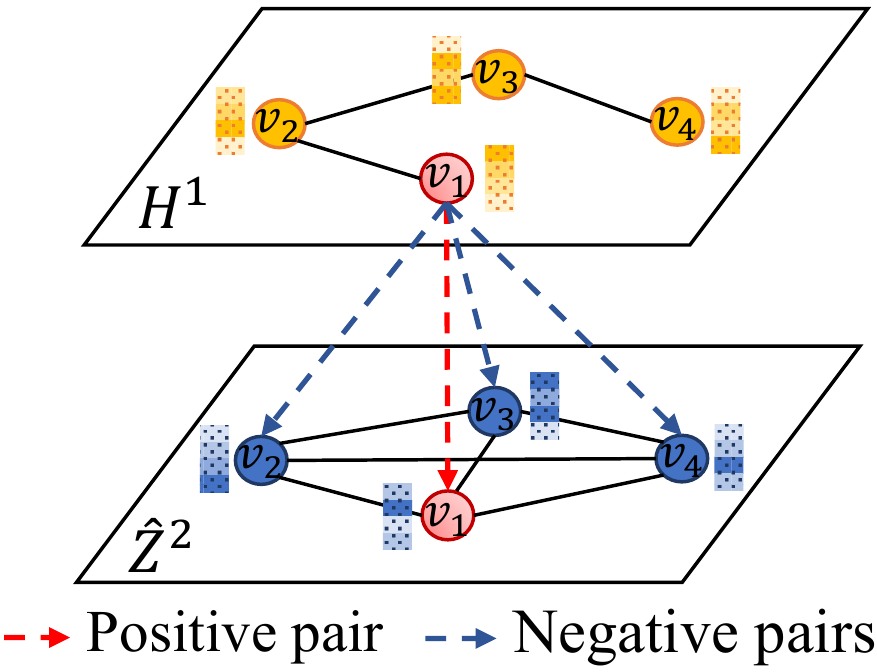}
\label{fig:picture002}
}
\subfigure[Cross-view contrastiveness.]{
\includegraphics[height = 3cm]{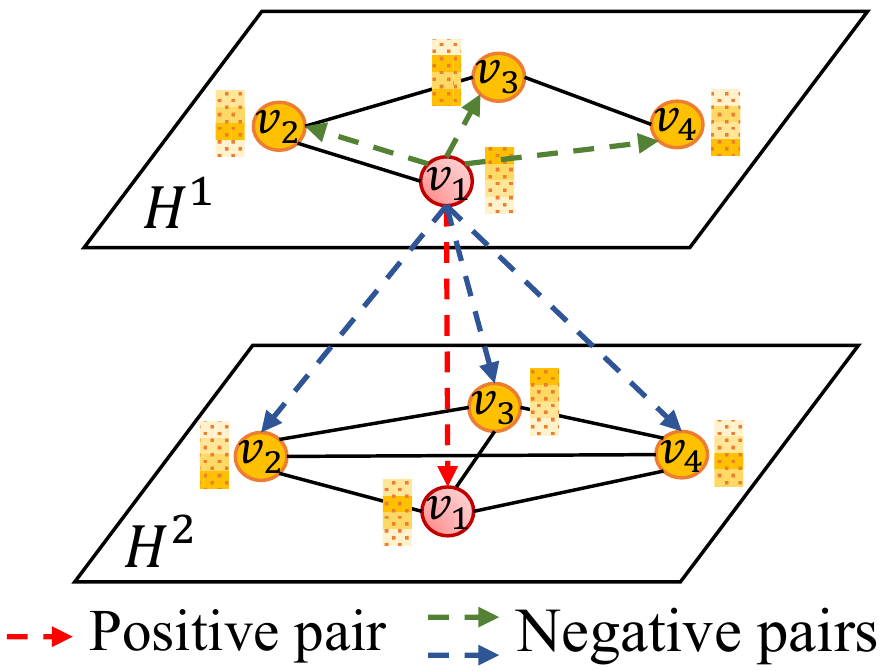}
\label{fig:picture003}
}
\caption{Cross-network contrastive learning is based on pairs from two different representations in the online and target network. Differently, cross-view contrastiveness discriminates the pair representations from two views in the same (i.e., online) network.}
\end{figure}

This contrastive learning process is illustrated in Figure \ref{fig:picture002}, where $H^1=q_{\theta}(Z^1)$ and $\hat{Z^2}$ denotes the representation of $\Tilde{\mathcal{G}_1}$ and $\Tilde{\mathcal{G}_2}$ from two different networks. Specifically, we use $Z^1=p_{\theta}\left(g_{\theta}(\Tilde{X_1},\Tilde{A_1})\right)$ and $\hat{Z^2}=p_{\zeta}\left(g_{\zeta}(\Tilde{X_2},\Tilde{A_2})\right)$ to denote the output node embeddings of our online and target encoder for the view 1 and 2. The red dash line between two $v_1$ nodes represents a positive pair $(h^{1}_{v_1}, \hat{z}^{2}_{v_1})^+$ constructed based on $v_1$. 
The intuition behind is to pull closer the representations of the same node from different views across two networks to distill the knowledge from historical observations, as well as stabilizing the online encoder training. To facilitate this, our target network does not directly receive the gradient during the training. Instead, we update its parameters by leveraging the momentum updating mechanism: 
\begin{equation}
\zeta^{t}=m \cdot \zeta^{t-1} + (1-m) \cdot \theta^{t},
\label{eq: moment}
\end{equation}
where $m$, $\zeta$, and $\theta$ are momentum, target network parameters, and online network parameters, respectively.

To further explore the rich contrastive relations between node representations within $H^1$ and $\hat{Z}_2$, we construct extra negative samples to regularize the basic bootstrapping loss, which are the blue dash lines between the red anchor node and blue nodes in Figure \ref{fig:picture002}, i.e., $(h^{1}_{v_1}, \hat{z}^{2}_{v_j})^-$, where we aim to push away from each other. 
Thus, the aforementioned processes can be formulated with the following loss functions:
\begin{equation}
\mathcal{L}^1_{cn}(v_i) = -\log \frac{\exp(\text{sim}(h^{1}_{v_i}, \hat{z}^{2}_{v_i}))}{\sum_{j=1}^{N}\exp(\text{sim}(h^{1}_{v_i}, \hat{z}^{2}_{v_j}))},
\label{eq: lcn1}
\end{equation}
\begin{equation}
\mathcal{L}^2_{cn}(v_i) = -\log \frac{\exp(\text{sim}(h^{2}_{v_i}, \hat{z}^{1}_{v_i}))}{\sum_{j=1}^{N}\exp(\text{sim}(h^{2}_{v_i}, \hat{z}^{1}_{v_j}))}.
\label{eq: lcn2}
\end{equation}

In above formulas, $\mathcal{L}^1_{cn}$ and $\mathcal{L}^2_{cn}$ are two symmetric losses, which represent the multi-scale cross-network contrastiveness on different views. Besides, $h^{1}_{v_i} \in H^1$, $h^{2}_{v_i} \in H^2$, $\hat{z}^{1}_{v_i} \in \hat{Z}^1$, $\hat{z}^{2}_{v_i} \in \hat{Z}^2$, and $\text{sim}(\cdot)$ denotes the cosine similarity.

Finally, by combining above two losses, we have our final cross-network contrastive objective function defined below:
\begin{equation}
\mathcal{L}_{cn} = \frac{1}{2N}\sum_{i = 1}^{N}\left(\mathcal{L}^1_{cn}(v_i) + \mathcal{L}^2_{cn}(v_i)\right).
\label{eq: lcn}
\end{equation}

\subsection{Cross-View Contrastive Learning}
Apart from the contrastive relations across two networks, the ties between two views within the online network have not been considered yet, which acts as a strong regularization to enhance the learning ability of our method. We do not have to include such contrastive relations within the target network because it will not directly receive the gradient, and our end goal is to train $g_{\theta}$ within the online encoder. Figure \ref{fig:picture003} illustrates our cross-view contrastiveness design, which consists of two discrimination schemes from two perspectives, namely the intra- and inter-view contrastiveness. Similar to but different from GRACE~\cite{grace}, we apply such contrastive learning on the top of differently-augmented views, where we consider not only the local structural and attributive augmentations (e.g., edge modification and feature masking) but also the global topological information injected via the graph diffusion.

We start from the inter-view contrastiveness, which pulls closer the representations of the same nodes in two augmented views while pushing other nodes away, as depicted by the red and blue dash lines in Figure \ref{fig:picture003}. In this case, we define our positive and negative pairs as $(h^{1}_{v_1}, h^{2}_{v_1})^+$ and $(h^{1}_{v_1}, h^{2}_{v_j})^-$. Similar to the objective functions used in the previous section, the inter-view contrastive loss $\mathcal{L}_{inter}$ for view 1 can be formulated as:
\begin{equation}
\mathcal{L}_{inter}^1(v_i) = -\log \frac{\exp(\text{sim}(h^{1}_{v_i}, h^{2}_{v_i}))}{\sum_{j=1}^{N}\exp(\text{sim}(h^{1}_{v_i}, h^{2}_{v_j}))}.
\label{eq: inter1}
\end{equation}

We can obtain $\mathcal{L}_{inter}^2(v_i)$ in similar way for view 2.
On the other hand, as the red and green dash lines shown in Figure \ref{fig:picture003}, the intra-view contrastiveness regards all nodes except the anchor node (i.e., $v_1$) as negatives within a particular view, denoted as $(h^{1}_{v_1}, h^{1}_{v_j})^-$. Thus, the intra-view contrastive loss $\mathcal{L}_{intra}$ for view 1 can be constructed below, which shares the same positive pairs with our inter-view contrastive losses:

\begin{equation}
\small
\begin{aligned}
\mathcal{L}_{intra}^1(v_i) &=
-\log \frac{\exp(\text{sim}(h^{1}_{v_i}, h^{2}_{v_i}))}{\exp(\text{sim}(h^{1}_{v_i}, h^{2}_{v_i})) + \Phi}, \\
\Phi &= \sum_{j=1}^{N}\mathds{1}_{i \neq j}\exp(\text{sim}(h^{1}_{v_i}, h^{1}_{v_j})),
\end{aligned}
\label{eq: intra1}
\end{equation}
where $\Phi$ denotes the accumulated similarity of negative pairs in the intra-view contrastive learning. Similarly, we can calculate $\mathcal{L}_{intra}^2(v_i)$ for view 2.

By combining the inter- and  intra-view contrastiveness of both views, we have our cross-view contrastive objective function $\mathcal{L}_{cv}$ formulated below:
\begin{equation}
\mathcal{L}_{cv} = \frac{1}{2N}\sum_{i = 1}^N\left(\mathcal{L}_{cv}^1(v_i) + \mathcal{L}_{cv}^2(v_i)\right),
\label{eq: lcv}
\end{equation}
where $\mathcal{L}_{cv}^1(v_i)$ and $\mathcal{L}_{cv}^2(v_i)$ are two symmetric losses that represent the multi-scale cross-view contrasitveness on the two views:
\begin{equation}
\mathcal{L}_{cv}^{k}(v_i) = \mathcal{L}_{intra}^k(v_i) + \mathcal{L}_{inter}^k(v_i), ~~ k \in \{1,2\}.
\label{eq: lcvk}
\end{equation}

\subsection{Model Training}
To train our model end-to-end and learn node representations for downstream tasks, we jointly leverage both the cross-view and cross-network contrastive loss. Specifically, the overall objective function is defined as:
\begin{equation}
\mathcal{L} = \beta\mathcal{L}_{cv} + (1 - \beta)\mathcal{L}_{cn},
\label{eq: finalobj}
\end{equation}
where we aim to minimize $\mathcal{L}$ during the optimization, and $\beta$ is a balance factor.
During the inference, we aggregate the representations generated by the online graph encoder $g_{\theta}$, taking both the graph adjacency and diffusion matrices as inputs: $\tilde{H} = H^1 + H^2 \in \mathbb{R}^{N \times D'}$, for downstream tasks.

\begin{table}[t]
\small
	\centering
	\begin{tabular}{@{}lcccc@{}}
		\toprule
		\textbf{Dataset} &  \textbf{Nodes} & \textbf{Edges} & \textbf{Features} & \textbf{Classes} \\ \midrule
		\textbf{Cora}              & 2,708    & 5,429        & 1,433            & 7                \\
		\textbf{CiteSeer}         & 3,327    & 4,732        & 3,703            & 6                \\
		\textbf{PubMed}           & 19,717   & 44,338      & 500               & 3                \\
		\textbf{Amazon Photo}    & 7,650     & 119,081       & 745              & 8                \\
		\textbf{Coauthor CS}     & 18,333     & 81,894       & 6,805              & 15            \\
 \bottomrule
	\end{tabular}
		\caption{The statistics of the datasets.}
	\label{tab:dataset}
\end{table}

\section{Experiment}
To evaluate the effectiveness of MERIT on self-supervised node representation learning, we conduct extensive experiments on five widely used benchmark datasets, including Cora, 
CiteSeer
, PubMed, 
Amazon Photo~\cite{shchur2018pitfalls}, and Coauthor CS~\cite{shchur2018pitfalls}. The dataset statistics are summarized in Table \ref{tab:dataset}.

\begin{table*}[htbp]
	\small
	\centering
	\begin{tabular}{llccccc}
		\toprule
		Information Used & Method & Cora & CiteSeer &PubMed & Amazon Photo & Coauthor CS\\
		\midrule
		\textbf{A, Y} & LP & 68.0 & 45.3  & 63.0 & 67.8$\pm{0.0}$  & 74.3 $\pm{0.0}$ \\
		\midrule
		\textbf{X, A, Y} &Chebyshev & 81.2 & 69.8  & 74.4 & 74.3$\pm{0.0}$  & 91.5 $\pm{0.0}$ \\
			\textbf{X, A, Y} &GCN & 81.5 & 70.3  & 79.0 & 87.3$\pm{1.0}$  & 91.8 $\pm{0.1}$ \\
			\textbf{X, A, Y} &GAT & 83.0 $\pm{0.7}$ & 72.5 $\pm{0.7}$  & 79.0 $\pm{0.3}$ & 86.2 $\pm{1.5}$  & 90.5 $\pm{0.7}$ \\
			\textbf{X, A, Y} &SGC & 81.0 $\pm{0.0}$ & 71.9 $\pm{0.1}$  & 78.9 $\pm{0.0}$ & 86.4 
		$\pm{0.0}$  & 91.0 $\pm{0.0}$ \\
		\midrule
			\textbf{X, A}	& DGI & 81.7 $\pm{0.6}$ & 71.5 $\pm{0.7}$  & 77.3 $\pm{0.6}$ & 83.1 $\pm{0.5}$  & 90.0 $\pm{0.3}$ \\
			\textbf{X, A}	&GMI & 82.7 $\pm{0.2}$ & 73.0 $\pm{0.3}$  & \textbf{80.1} $\pm{\bm{0.2}}$ & 85.1 $\pm{0.1}$  & 91.0 $\pm{0.0}$ \\
			\textbf{X, A}	&MVGRL & 82.9 $\pm{0.7}$ & 72.6 $\pm{0.7}$  & 79.4 $\pm{0.3}$ & 87.3 $\pm{0.3}$  & 91.3 $\pm{0.1}$ \\
			\textbf{X, A}	&GRACE & 80.0 $\pm{0.4}$ & 71.7 $\pm{0.6}$  & 79.5 $\pm{1.1}$ & 81.8 $\pm{1.0}$  & 90.1 $\pm{0.8}$ \\
		\midrule
			\textbf{X, A}	& 	MERIT & \textbf{83.1} $\pm{\bm{0.6}}$ & \textbf{74.0} $\pm{\bm{0.7}}$  & \textbf{80.1} $\pm{\bm{0.4}}$ & \textbf{87.4} $\pm{\bm{0.2}}$  & \textbf{92.4} $\pm{\bm{0.4}}$ \\
		\bottomrule
	\end{tabular}
	\caption{Classification accuracies on five benchmark datasets. $\textbf{X}$, $\textbf{A}$, and $\textbf{Y}$ indicate the node feature, adjacency matrix, and label information exploited by each algorithm, respectively. Some results without standard deviations are directly taken from~\protect\cite{mvgrl}.}
\label{tab: classification results}
\end{table*}

\subsection{Experimental Settings}
For simplicity, 
we adopt a 1-layer GCN~\cite{gcn} as our backbone graph encoders (i.e., $g_{\theta}$ and $g_{\zeta}$). 
For model tuning, we perform the grid search on primary hyper-parameters over certain ranges. The latent dimension of graph encoders, projectors, and the predictor is fixed to 512. 
We tune momentum $m$ and augmentation ratio $P$ between 0 and 1.
To balance the effect of 
two contrastive schemes, we tune $\beta$ within $\{0.2, 0.4, 0.6, 0.8\}$.

To evaluate the trained graph encoder, we adopt a linear evaluation protocol by training a separate logistic regression classifier on top of the learned node representations.
For Cora, Citeseer and PubMed, we follow the same data splits as in \cite{yang2016revisiting}. 
For Amazon Photo and Coauthor CS, we include 30 randomly-selected nodes per class to construct the training and validation set, while using the remaining nodes as the testing set. We repeat the experiments in Table \ref{tab: classification results} and \ref{tab: ablation} ten times and report the average accuracy with the standard deviation. 

\subsection{ Classification Results}
We choose node classification as our downstream task and compare MERIT with five supervised methods and four state-of-the-arts graph contrastive learning models. For supervised baselines, we select LP~\cite{zhu2003semi}, ChebConv~\cite{chebyshev}, GCN~\cite{gcn}, GAT~\cite{gat}, and SGC~\cite{sgc}. DGI~\cite{dgi}, MVGRL~\cite{mvgrl}, GMI~\cite{gmi}, and GRACE~\cite{grace} are chosen as the self-supervised competitors. 
We report the overall classification results in Table~\ref{tab: classification results} and highlight the best performance in bold.

We can observe from the table that MERIT achieves the best classification accuracy on all five datasets, surpassing not only the self-supervised but also supervised methods (except a draw with GMI on PubMed). This result can be attributed to two key components in our framework: (1). Different from the compared GCL methods, we introduce a more expressive Siamese architecture to help the graph encoder distill the knowledge from historical representations and alleviate the reliance on negative samples. (2). To further realise the potential of our model, we introduce multiple contrastive routes within and across different views and networks, which not only provide a stronger regularization to our bootstrapping objective but also enrich the self-supervision signals during the optimization. 

By comparing the best performances of selected supervised and self-supervised baselines, we observe that contrastive learning-based models have achieved similar or even better classification accuracy in more than one datasets, which demonstrates the effectiveness of mining rich multi-scale contrastive relations in graphs. However, there still exist performance gaps between state-of-the-art graph contrastive learning and supervised methods in Cora and Coauthor CS.

\subsection{Parameter Sensitivity Study}
\begin{figure}[bp]
\centering
\includegraphics[width=.35\textwidth, height=.23\textwidth]{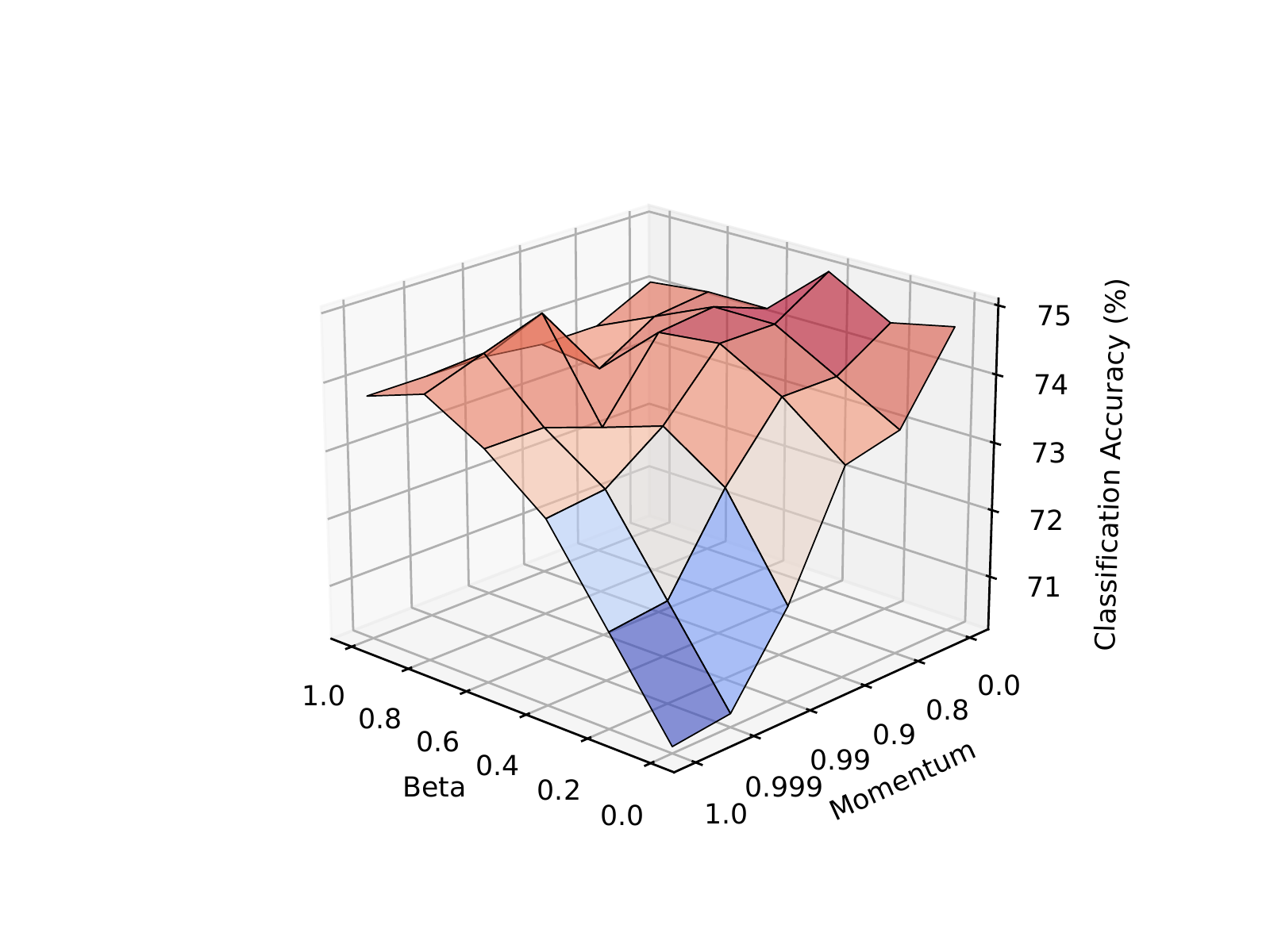}
\label{fig5}
\caption{Classification accuracies of MERIT on CiteSeer with different $\beta$ and $m$. A warmer color denotes a higher accuracy.}
\label{fig:para}
\end{figure}

\paragraph{Balance factor $\beta$ and momentum $m$.} 
We study two important hyper-parameters, the balance factor $\beta$ in our final objective function (i.e., Equation~\ref{eq: finalobj}) and the momentum term $m$ in Equation~\ref{eq: moment}.
In Figure~\ref{fig:para}, for a fixed momentum value, we observe that a $\beta$ value between 0.4 and 0.6 typically produces the best accuracies, which confirms our conjecture that the two proposed contrastive losses can regularize each other and achieve better results than only optimizing one of them (i.e.,\ $\beta=0$ or $\beta=1$), where we give a detailed analyze in our ablation study. On the other hand, for a certain $\beta$ value, we find that $m=1$ usually gives a poor performance, comparing with other values between 0 and 0.999 as recommended in BYOL~\cite{byol} and SimSiam~\cite{simsiam}. We conjecture that making the parameters stale in the target network may hinder the process of knowledge distillation and thus disturb the model optimization. Another interesting finding here is that our model even performs better when $m=0$. Our hypothesis is that the real effective components in bootstrapping are the predictor and stop gradient in our framework, which we leave to research in 
further work.

\begin{figure}[tbp]
\subfigure[Study on structural and attributive graph augmentations.]{
\label{Fig.augsub.1}
\includegraphics[width=0.23\textwidth]{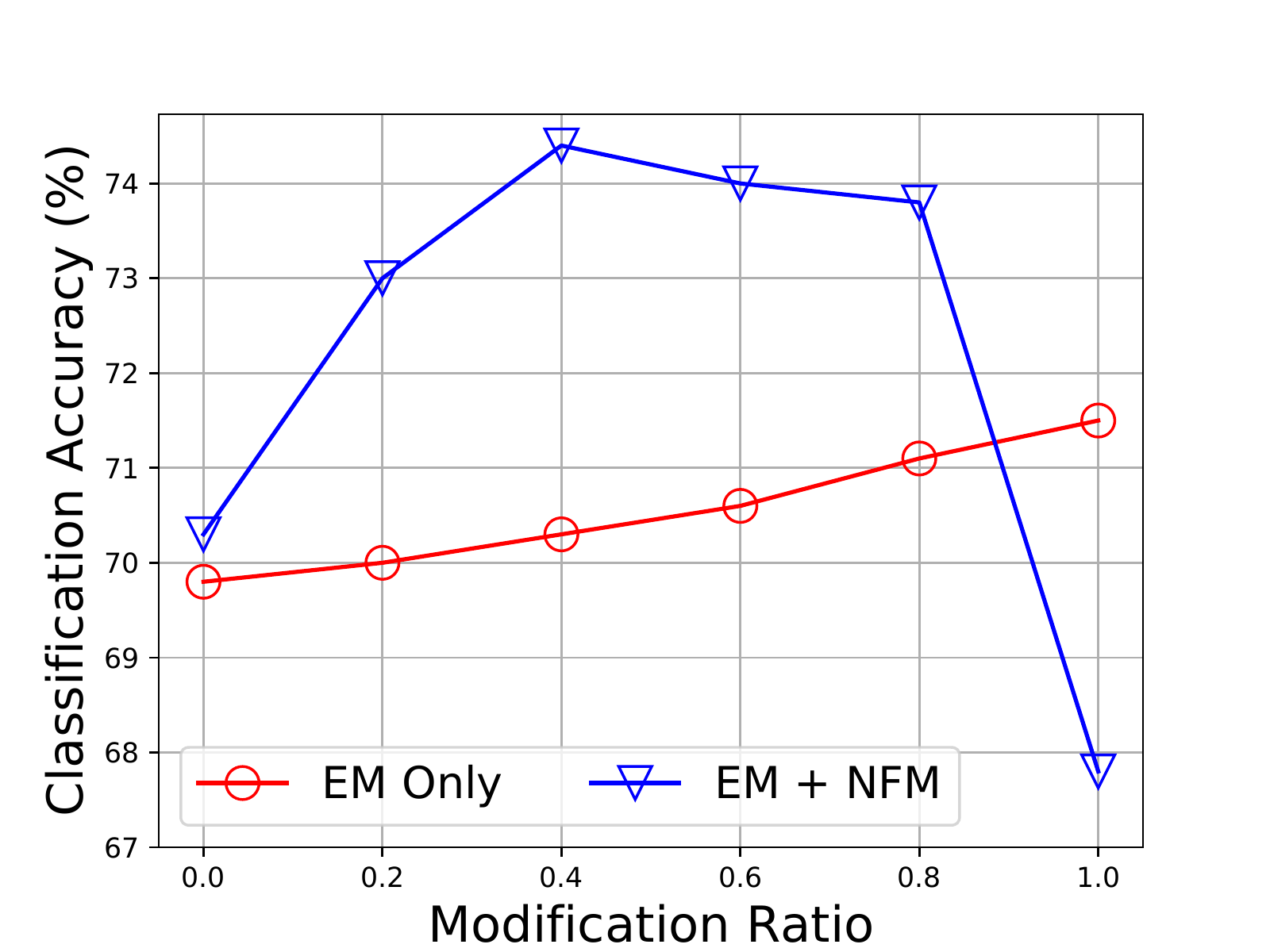}}
\subfigure[Effect of graph diffusion on performance.]{
\label{Fig.augsub.2}
\includegraphics[width=0.23\textwidth]{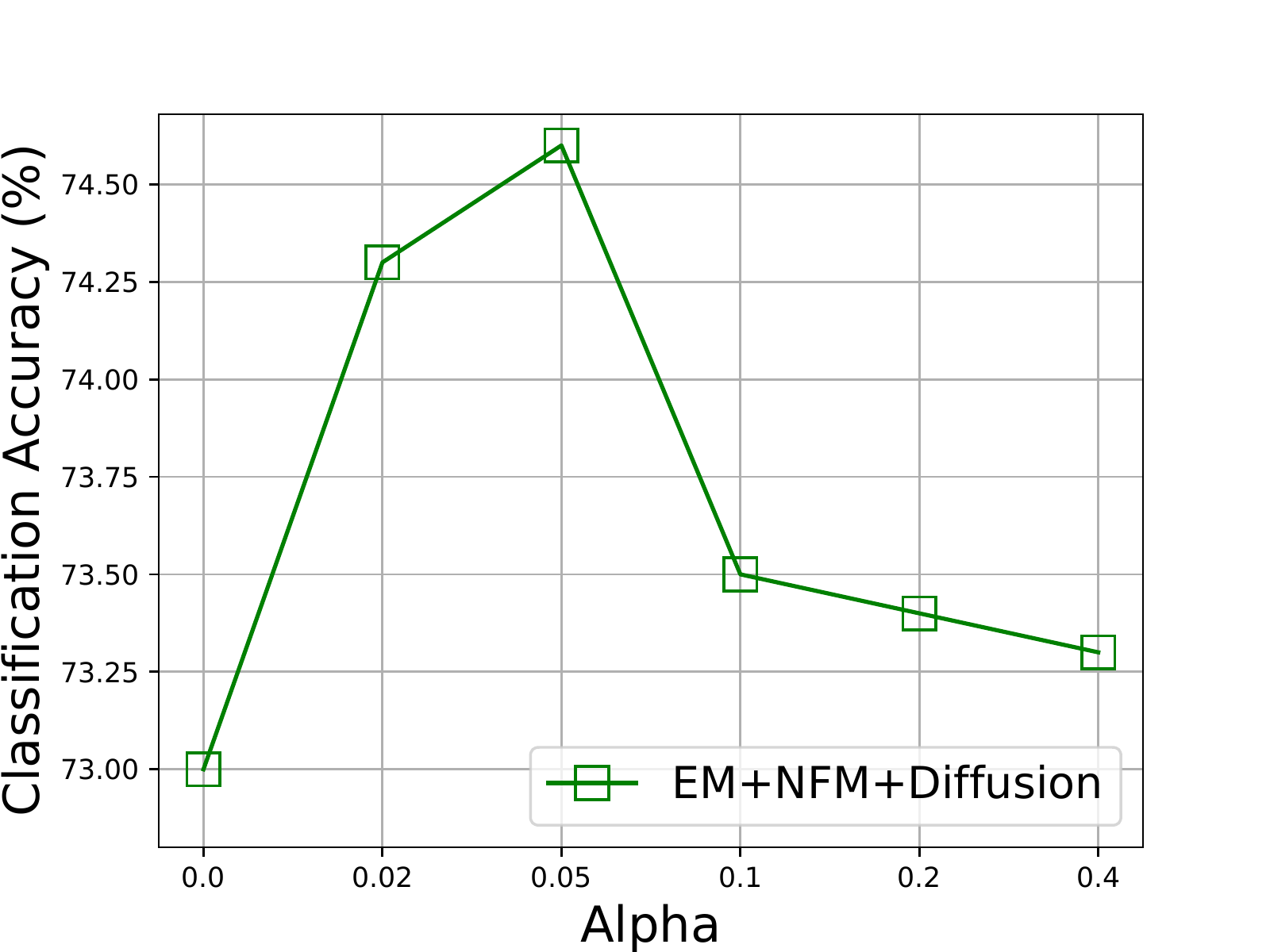}}
\caption{Classification accuracies versus graph augmentation in varying types and degrees.}
\label{fig: aug}
\end{figure}

\begin{table}[tbp]
	\small
	\centering
	\begin{tabular}{lrr}
		\toprule
		\textbf{Method} & \textbf{CiteSeer} & \textbf{Amazon Photo}\\
		\midrule
		MERIT & 74.0 $\pm{0.7}$ & 87.4  $\pm{0.2}$\\
		MERIT w/o cross-network & 73.8  $\pm{0.4}$ & 87.0 $\pm{0.1}$  \\
		MERIT w/o cross-view & 73.6  $\pm{0.4}$ & 87.1 $\pm{0.3}$ \\
		\bottomrule
	\end{tabular}
	\caption{Ablation study on CiteSeer and Amazon Photo}
	\label{tab: ablation}
\end{table}

\paragraph{Effect of augmentation.}
Apart from $m$ and $\beta$, augmentation plays a critical role in contrastive learning. Comparing the red and blue lines in Figure~\ref{Fig.augsub.1}, we observe that jointly considering the structure and attributive augmentations gives the best performance. When applying the edge modification only, we surprisingly find that a higher modification ratio gives a better performance. We conjecture that this is caused by the nature of contrastive learning, which requires a more challenging pretext task to achieve competitive performance. When considering multi-scale augmentations, e.g., combining edge and node feature modification, overly increasing the modification ratio may distort the underlying topological and attributive information, thus leading to significant performance degrade. The effectiveness of graph diffusion can be observed in Figure \ref{Fig.augsub.2}, where $\alpha=0.05$ gives the best performance in our experiments. When removing this module (i.e., $\alpha=0$), the performance decreases dramatically, which confirms our hypothesis that injecting global information further boosts the expressive ability of our model.

\subsection{Ablation Study}
To validate the effectiveness of the two contrastive components, 
we conduct experiments on Citesser and Amazon Photo for two MERIT variants, each of which has one of the key components removed. The result is presented in Table \ref{tab: ablation}. 
Here we use MERIT w/o cross-network and MERIT w/o cross-view to denote the ablated model with cross-network loss $\mathcal{L}_{cn}$ or cross-view loss $\mathcal{L}_{cv}$ being masked. From Table \ref{tab: ablation}, we can find that our model performance would degrade without one of the key components on the two datasets, 
which demonstrates the effectiveness of our two contrastive schemes.
Specifically, our proposed model can boost MERIT w/o cross-view with 0.4\% and 0.3\% improvement, and MERIT w/o cross-network with 0.2\% and 0.4\% improvement for CiteSeer and Amazon Photo, respectively. This improvement can be attributed to our comprehensive multi-scale contrastive learning scheme, which takes the advantage of both single- and multiple-network contrastiveness.

\begin{figure}[tbp]
\centering  
\subfigure[GCN]{
\label{Fig.tsne.sub.1}
\includegraphics[width=0.15\textwidth]{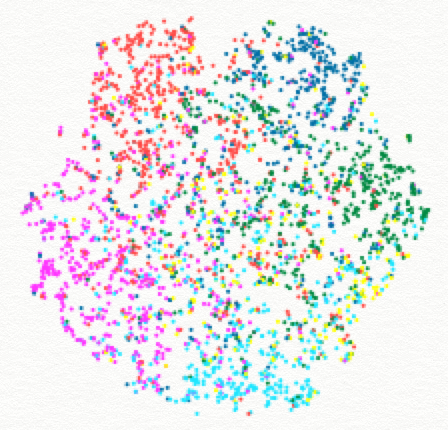}}
\subfigure[DGI]{
\label{Fig.tsne.sub.2}
\includegraphics[width=0.15\textwidth]{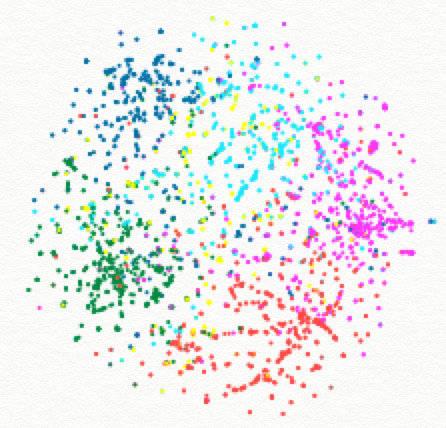}}
\subfigure[MERIT]{
\label{Fig.tnse.sub.3}
\includegraphics[width=0.15\textwidth]{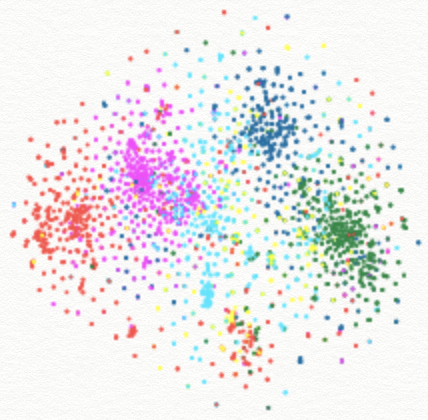}}
\caption{t-SNE embeddings of nodes in the CiteSeer dataset.}
\label{fig: tsne}
\end{figure}

\paragraph{Visualisation.} 
To show the superiority of our model, we visualize the node embeddings of CiteSeer calculated by GCN, DGI, and MERIT via the t-SNE algorithm, in which node colors denote different classes. 
In Figure~\ref{fig: tsne}, MERIT's 2D projection presents a clearer separation, which indicates
that 
our approach benefits the 
graph encoder to extract more expressive node representations for downstream tasks.

\section{Conclusion}
In this paper, we present a novel approach towards self-supervised graph representation learning.
By leveraging the backbone Siamese GNNs, we design a cross-network contrastiveness to distill the knowledge from historical representations to guide and stabilize the training of online graph encoder. To further enrich the self-supervision signals, we introduce another cross-view contrastive objective on multiple scales to regularize the bootstrapping scheme in the cross-network contrastiveness. Experimental results demonstrate the superiority and the effectiveness of our method.

\bibliographystyle{named}
\bibliography{ijcai21}

\end{document}